\documentclass[acmsmall]{acmart}

\usepackage{multirow}
\usepackage{amsfonts}
\usepackage{color}
\usepackage{subfigure}
\usepackage{adjustbox,lipsum}
\usepackage{array}
\usepackage{algorithm}
\usepackage{algorithmic}
\usepackage{rotating}

\definecolor{orange}{rgb}{1,0.5,0}
\definecolor{grey}{rgb}{0.5,0.5,0.5}
\definecolor{new_blue}{rgb}{0.3,0.5,0.8}

\newcommand{\PreserveBackslash}[1]{\let\temp=\\#1\let\\=\temp}
\newcolumntype{C}[1]{>{\PreserveBackslash\centering}p{#1}}
\newcolumntype{R}[1]{>{\PreserveBackslash\raggedleft}p{#1}}
\newcolumntype{L}[1]{>{\PreserveBackslash\raggedright}p{#1}}

\newcommand{\etal}{{\em et al.\,}}       
\newcommand{\eg}{{\em e.g.}}           
\newcommand{\ie}{{\em i.e.}}           

\newcommand{\bl}{\textcolor{blue}}

\AtBeginDocument{%
  \providecommand\BibTeX{{%
    \normalfont B\kern-0.5em{\scshape i\kern-0.25em b}\kern-0.8em\TeX}}}

\setcopyright{acmcopyright}
\acmJournal{TOMM}
\acmYear{2023} \acmVolume{1} \acmNumber{1} \acmArticle{1} \acmMonth{1} \acmPrice{15.00}\acmDOI{10.1145/3419439}

\begin{document}

\title{DoubleAUG: Single-domain Generalized Object Detector in Urban via Color Perturbation and Dual-style Memory}

\author{Lei Qi, Peng Dong, Tan Xiong, Hui Xue and Xin Geng}
\affiliation{%
  \institution{Key Laboratory of New Generation Artificial Intelligence Technology and Its Interdisciplinary Applications (Southeast University),
Ministry of Education}
  \city{Nanjing}
  \country{China}}
\email{qilei@seu.edu.cn, dongpeng@seu.edu.cn, xiongtan@seu.edu.cn, hxue@seu.edu.cn and xgeng@seu.edu.cn}





\thanks{The work is supported by NSFC Program (Grants No. 62206052, 62125602, 62076063) and Jiangsu Natural Science Foundation Project (Grant No. BK20210224). Corresponding author: Xin Geng.}



\begin{abstract}
Object detection in urban scenarios is crucial for autonomous driving in intelligent traffic systems. However, unlike conventional object detection tasks, urban-scene images vary greatly in style. For example, images taken on sunny days differ significantly from those taken on rainy days. Therefore, models trained on sunny day images may not generalize well to rainy day images. In this paper, we aim to solve the single-domain generalizable object detection task in urban scenarios, meaning that a model trained on images from one weather condition should be able to perform well on images from any other weather conditions.
To address this challenge, we propose a novel Double AUGmentation (DoubleAUG) method that includes image- and feature-level augmentation schemes. In the image-level augmentation, we consider the variation in color information across different weather conditions and propose a Color Perturbation (CP) method that randomly exchanges the RGB channels to generate various images. In the feature-level augmentation, we propose to utilize a Dual-Style Memory (DSM) to explore the diverse style information on the entire dataset, further enhancing the model's generalization capability.
Extensive experiments demonstrate that our proposed method outperforms state-of-the-art methods. Furthermore, ablation studies confirm the effectiveness of each module in our proposed method. Moreover, our method is plug-and-play and can be integrated into existing methods to further improve model performance.
\end{abstract}




\maketitle
\section{Introduction}
Object detection is a critical technology for intelligent traffic systems~\cite{DBLP:conf/iccv/Girshick15,DBLP:journals/tits/HuPSHP16}. In urban scenarios, accurately locating objects is essential for improving the security of autonomous driving systems. With the success of deep learning~\cite{aversa2020deep}, typical object detection tasks have made significant breakthroughs in the computer vision community. This development has also effectively supported object detection in urban scenarios.

\begin{figure}
  \centering
  \subfigure[Diverse-Weather]{
    \includegraphics[width=6.61cm]{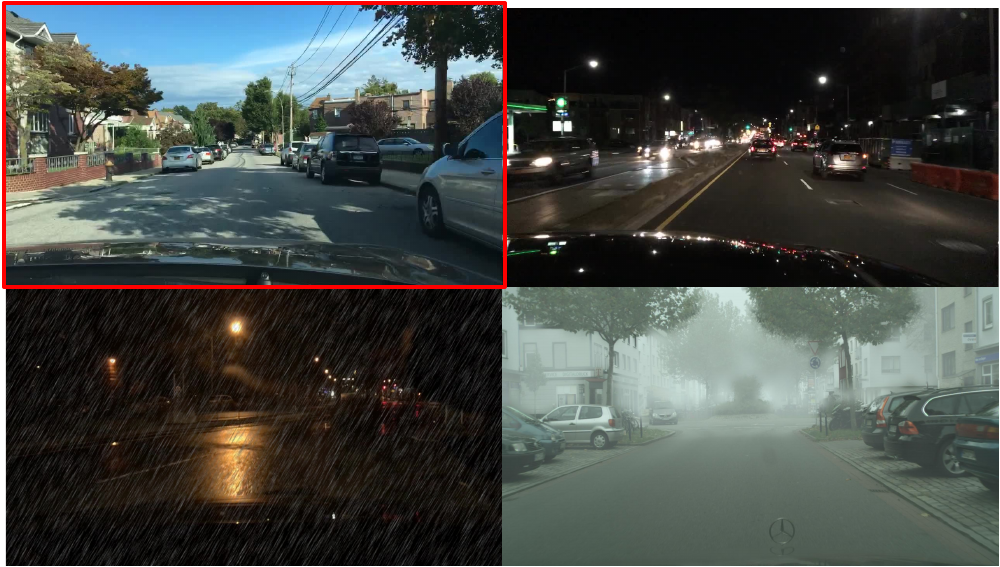}}
  \hfill
  \subfigure[SIM10k2Cityscapes]{
        \includegraphics[width=6.61cm]{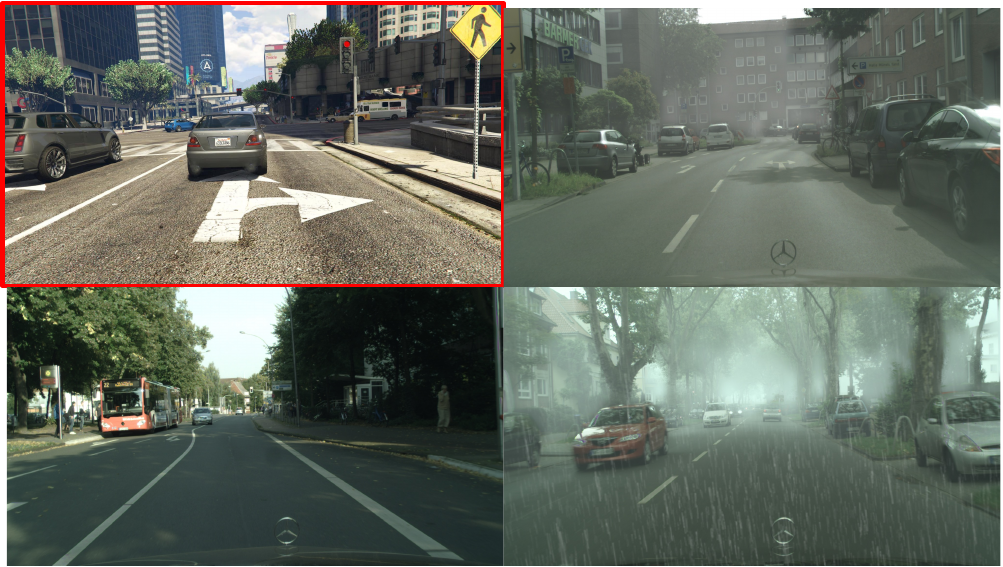}
        }
  \caption{Images from two tasks. In both figures, all images are from different domains. The image with the red box represents the image from the training set, while the other images belong to the testing domain. As illustrated, there is an evident difference in style between the training and testing samples.} 
  \label{fig03}
\end{figure}

\begin{figure}
  \centering
  \subfigure[Diverse-Weather]{
    \includegraphics[width=5.61cm]{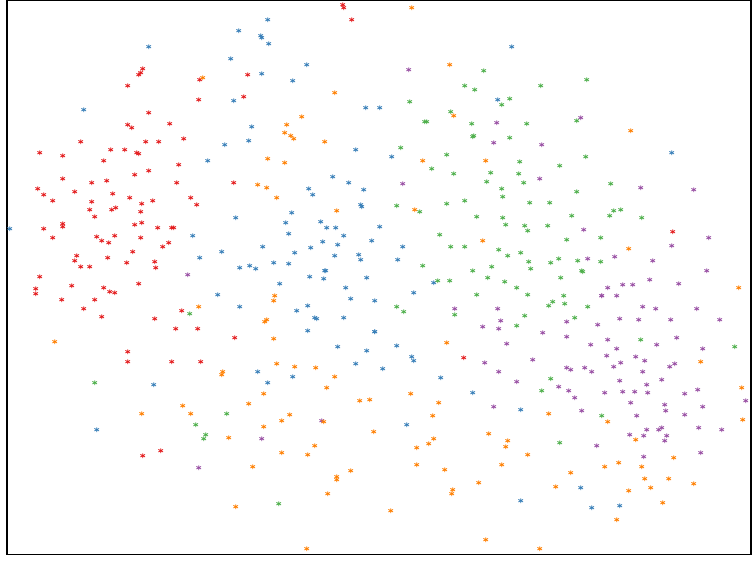}}
  \subfigure[SIM10k2Cityscapes]{
        \includegraphics[width=5.61cm]{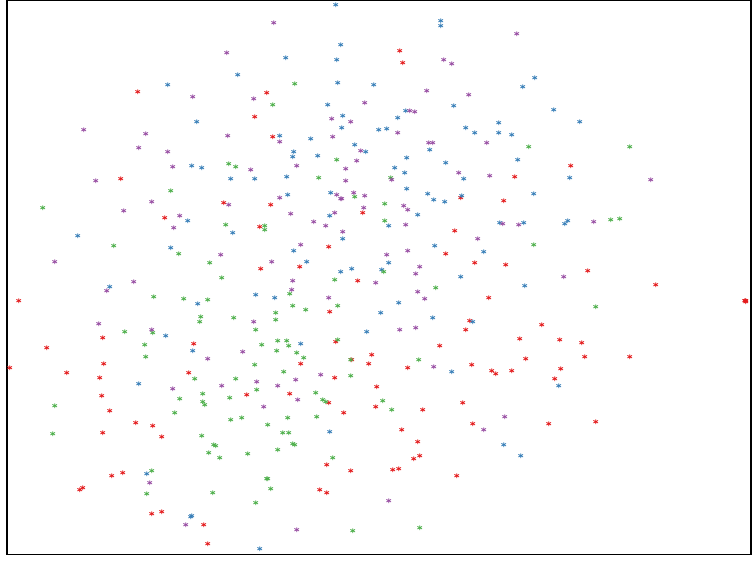}
        }
  \caption{Visualization of the data distribution in two tasks using t-SNE~\cite{van2008visualizing}. The feature is extracted using the ResNet~\cite{DBLP:conf/cvpr/HeZRS16} pre-trained on ImageNet~\cite{deng2009imagenet}. Note that different colors denote different domains. To generate the figure, we randomly select 100 samples from each domain.} 
  \label{fig06}
\end{figure}

In urban scenarios, the image style often varies with weather and time. For example, images collected during the daytime differ from those collected at night, leading to different data distributions across scenarios. Fig.~\ref{fig03} displays samples from two tasks, each belonging to a different domain, highlighting the obvious discrepancy between images. Besides, in Fig.~\ref{fig06}, we also visualize the data distribution from different scenarios using t-SNE~\cite{van2008visualizing}, showing that images from different scenarios (\ie, domains) occupy different spaces. If we train a object detector using images collected during the daytime, it may not generalize well to images collected in other scenarios. Therefore, directly applying the typical object detection method into urban scenarios requires collecting images from diverse scenarios. However, in real-world applications, since unseen scenarios are unpredictable, it is extremely challenging to collect images from all possible scenarios.

Additionally, even if we are able to collect images from all possible scenarios, labeling them would still be a costly and time-consuming task. While some unsupervised domain adaptation methods have been proposed to address this issue~\cite{DBLP:journals/tomccap/WuLSZ22,DBLP:journals/tomccap/XuSDWXH22,zhang2023multi}, they still require collecting data from the target scenario. However, as aforementioned, many target scenarios are unknown in real-world applications, making it challenging to collect sufficient data for training. In this paper, we focus on solving the single-domain generalization object detection (Single-DGOD) task in the urban scenario~\cite{WuAming_2022_CVPR}, which involves training the model on one domain (scenario) and testing it on other unseen domains, without the need for extensive data collection.

To address the single-domain generalizable object detection task in the urban scenario, most existing methods rely on adversarial learning-based augmentation techniques~\cite{zhou2020deep,DBLP:conf/cvpr/FanWKYGZ21}, which are designed primarily for image classification tasks. In the meanwhile, for the single-domain object detection (Single-DGOD) task, different approaches have been proposed. For instance, Wu  \etal \cite{WuAming_2022_CVPR} introduce the self-distillation method, which disentangles domain-specific and domain-invariant representations without the supervision of domain-related annotations. Wang \etal \cite{DBLP:conf/iccv/0066HLY0GD21} propose a self-supervised approach based on finding the most different object proposals in adjacent video frames and then cycling back to itself. However, in this paper, the proposed DoubleAUG method aims to solve the single-domain generalizable object detection task through a novel combination of image-level and feature-level augmentation schemes, rather than relying on adversarial learning or self-supervision techniques.

In this study, we propose a new method called Double AUGmentation (DoubleAUG) that incorporates both image- and feature-level augmentation to solve the single-domain object detection task in urban scenarios. Our image-level augmentation, Color Perturbation (CP), perturbs color information to combat overfitting caused by varying color characteristics across different scenarios. Our feature-level augmentation, Dual-Style Memory (DSM), leverages style information from the entire training set to increase diversity by switching styles of objects and backgrounds. We conduct numerous experiments to demonstrate the effectiveness of our approach and confirm the effectiveness of each module. Additionally, we analyze the efficacy of DoubleAUG using existing domain generalization theory. Our contributions include the development of DoubleAUG, an effective method for single-domain object detection, and the confirmation of the effectiveness of CP and DSM. The details are as follows:

  \begin{itemize}
    \item We present a simple but powerful double data augmentation method called DoubleAUG. This approach can generate a variety of color perturbations and utilize style information from the entire training set, thereby improving the robustness and efficacy of the model in detecting objects in unseen domains. Moreover, our method is plug-and-play and can be integrated into existing methods to further improve model performance.
    \item 
    To implement DoubleAUG, we introduce two components: Color Perturbation, which disturbs the RGB channels to enhance the color information in the image-level space, and Dual-Style Memory, which mines diverse style information in the feature space. These two components work together to provide a comprehensive data augmentation strategy for improving the robustness of object detection models in unseen domains.
    \item 
    We conduct extensive evaluations of our approach on multiple standard benchmark datasets and demonstrate that our approach outperforms the state-of-the-art in terms of accuracy. Additionally, we perform ablation studies and further analysis to validate the effectiveness of our method. Furthermore, we also analyze the efficacy of the proposed method using the existing domain generalization theory.
  \end{itemize}

The structure of this paper is as follows. In Section \ref{s-related}, we provide a review of related work. Section \ref{s-framework} details our proposed method, including Color Perturbation and Dual-Style Memory. Section ~\ref{EDET} analyzes the efficacy of the proposed method using existing domain generalization theory. Experimental results and analysis are presented in Section \ref{s-experiment}, followed by the conclusion in Section \ref{s-conclusion}.

\section{Related work}\label{s-related}
In this section, we review some related work on object detection, domain adaptive object detection, and domain generalization. Detailed introductions will be given in the following parts.

\subsection{Object Detection}
Object detection is a fundamental task in computer vision, which has been extensively studied for several years. Following the lead of RCNN ~\cite{DBLP:conf/cvpr/GirshickDDM14,DBLP:journals/pami/RenHG017}, numerous object detection frameworks~\cite{DBLP:conf/iccv/Girshick15,DBLP:conf/iccv/HeGDG17,DBLP:journals/pami/RenHG017} based on convolutional networks have been developed in recent years, which have significantly pushed forward the state-of-the-art performance. Object detection models can be broadly classified into two types: one-stage and two-stage detection. One-stage object detection refers to a class of object detection methods that skip the region proposal stage of two-stage models and directly run detection over a dense sampling of locations. YOLO~\cite{DBLP:journals/corr/abs-1804-02767} outputs sparse detection results with high computation speed. Two-stage detectors generate region proposals for detection, for instance, Faster-RCNN~\cite{DBLP:journals/pami/RenHG017,DBLP:conf/cvpr/GirshickDDM14}, which introduces the Region Proposal Network (RPN) for proposal generation. FPN~\cite{DBLP:conf/cvpr/LinDGHHB17} employs multiple layers for the detection of different scales.

\subsection{Domain Adaptive Object Detection}

To address the domain shift issue in object detection, several unsupervised domain adaptive methods have been proposed, such as~\cite{ DBLP:conf/cvpr/Chen0SDG18,DBLP:conf/eccv/HsuTLY20,DBLP:conf/cvpr/HsuHTYTS019,DBLP:conf/cvpr/SaitoUHS19,DBLP:conf/cvpr/XuZJW20,DBLP:conf/cvpr/XuWNTZ20}. These methods aim to align the feature-level distributions between the source and target domains. For instance, Wu \etal~\cite{DBLP:conf/cvpr/WuCHWLMGWW022} propose a teacher-student framework to extract knowledge from labeled source domains and guide the student network to learn detectors in the unlabeled target domain. Some methods ~\cite{DBLP:conf/cvpr/ChenZD0D20,DBLP:journals/pami/WuHZY22} attempt to extract instance-invariant features to improve the generalization ability. Additionally, Li \etal~\cite{DBLP:conf/cvpr/LiLY22} introduce a framework that employs a Graph-embedded Semantic Completion module to complete mismatched semantics and model class-conditional distributions with graphs. Although these methods have demonstrated their effectiveness, they typically require access to both the source and target domain data during training, limiting their applicability to domain generalization tasks.

\subsection{Domain Generalization}

Domain generalization aims at extracting knowledge from one or multiple source domains so as to generalize well to unseen target domains~\cite{10.1145/3524136}. Some existing DG methods~\cite{DBLP:conf/cvpr/HararySASHAHAGK22,DBLP:conf/cvpr/TeneyALH22,DBLP:conf/eccv/YangWK22,DBLP:conf/cvpr/ZhangZLWSX22,DBLP:conf/cvpr/ZhangLLJZ22} are proposed to minimize the difference among source domains for learning domain-invariant representations. Another popular way to address DG problems is domain augmentation~\cite{DBLP:conf/cvpr/KangLKK22,DBLP:conf/nips/VolpiNSDMS18,DBLP:conf/eccv/WangYLFH20,DBLP:conf/aaai/ZhouYHX20,DBLP:conf/eccv/ZhouYHX20}, which create samples from fictitious domains. Besides, other DG methods are proposed, such as learning strategies~\cite{DBLP:conf/cvpr/ChenL0LY22,DBLP:conf/aaai/LiYSH18,DBLP:conf/cvpr/WanSZY0GH022,DBLP:conf/eccv/Zhang0SG22} and so on~\cite{DBLP:conf/eccv/MengLCYSWZSXP22,DBLP:conf/eccv/MinPKPK22,DBLP:conf/cvpr/YaoBZZSCL022,DBLP:conf/cvpr/ZhangZX0SL22}. For example, in the training stage, Zhang \etal \cite{DBLP:conf/eccv/Zhang0SG22}  develop a multi-view regularized meta-learning algorithm that employs multiple optimization trajectories to produce a suitable optimization direction for model updating. Although current DG methods achieve the promising results, most of them use multi-domain data to train the model, which is unrealistic for some real-world applications.

\subsection{Single Domain Generalization}

Recently, a new task called single domain generalization has been proposed~\cite{DBLP:conf/nips/VolpiNSDMS18} , which aims to generalize a model trained on one source domain to any unseen domains. Most existing methods solve this task by employing data augmentation and feature normalization. For example, Volpi \etal~\cite{DBLP:conf/nips/VolpiNSDMS18} and Qiao \etal~\cite{DBLP:conf/cvpr/QiaoZP20} explore the use of adversarial mechanisms to solve this task, which helps promote large domain transportation in the input space. Wang \etal~\cite{DBLP:conf/iccv/WangLQHB21} aim to improve generalization by alternating diverse sample generation and discriminative style-invariant representation learning. Fan \etal~\cite{DBLP:conf/cvpr/FanWKYGZ21} propose a generic normalization approach, adaptive standardization and rescaling normalization, to improve generalization.

However, these methods cannot be directly applied to single domain generalization object detection (Single-DGOD). Therefore, some methods~\cite{DBLP:conf/iccv/0066HLY0GD21,DBLP:conf/cvpr/WuD22}  have attempted to solve this problem from different angles. Wang \etal~\cite{DBLP:conf/iccv/0066HLY0GD21} attempt to find the most different object proposals in adjacent frames in a video and then cycle back to itself for self-supervision. Wu \etal~\cite{DBLP:conf/cvpr/WuD22} present the cyclic-disentangled self-distillation method by disentangling domain-invariant representations from domain-specific representations without the supervision of domain-related annotations. In this paper, we propose a novel approach that aims to augment data at both the feature-level and image-level by exchanging the object and background information of different images in style and distorting the RGB channels of images to solve the Single-DGOD problem, respectively.


\begin{figure*}[t]
\centering
\includegraphics[width=14cm]{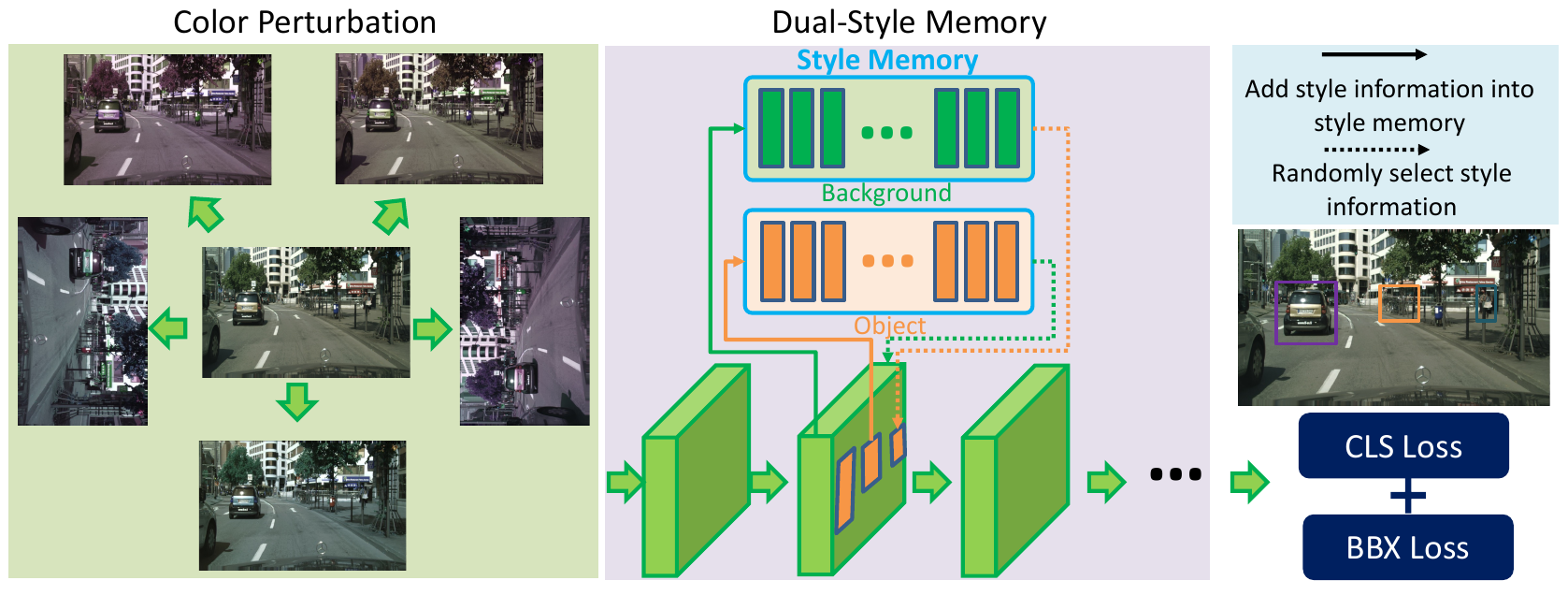}
\caption{An illustration of the proposed DoubleAUG, which consists of the image-level Color Perturbation (CP) and the feature-level Dual-Style Memory (DSM). It is worth noting that our method is plug-and-play and can be inserted into existing methods.}
\label{fig04}
\end{figure*}

\section{The proposed method}\label{s-framework}

In this paper, we propose a novel approach, called DoubleAUG, to enhance the generalization capacity of the model to the unseen domain, as illustrated in Fig.~\ref{fig04}. Specifically, we introduce color perturbations to conduct image augmentation, which effectively alleviates overfitting caused by color variations in the unseen domain. Additionally, we develop a style memory to explore and mine diverse style information from the entire training data, which further improves the model's generalization ability. The detailed description of the proposed method will be presented in the following section.

\subsection{Color Perturbation}
\begin{figure*}[t]
\centering
\includegraphics[width=12cm]{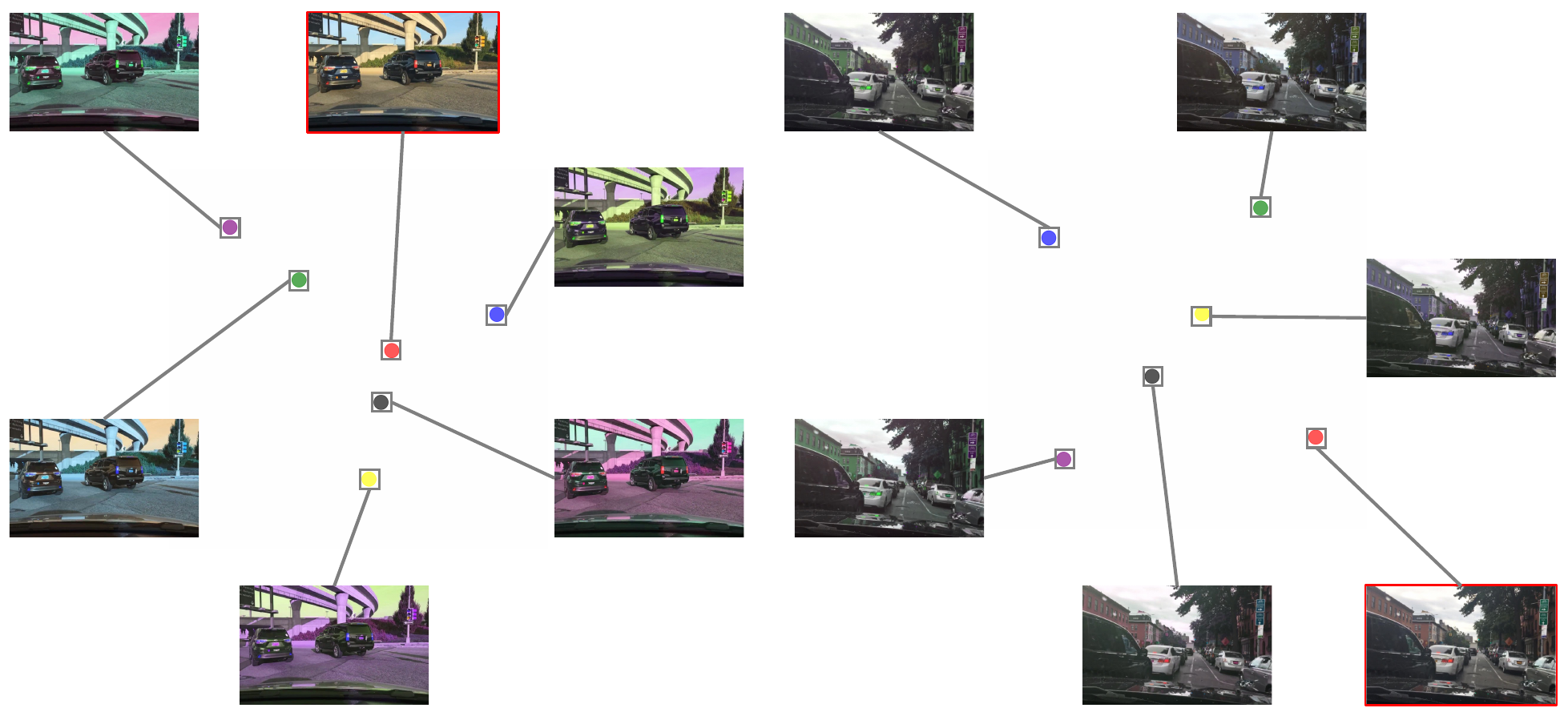}
\caption{Results with color perturbation. Note that the red box indicates the original image.}
\label{fig07}
\end{figure*}

In the single-domain urban-object detection task, the lack of diversity in the training set can lead to overfitting of the model to the training data. This is particularly true since the images in the dataset are collected at different times and in different weather conditions, resulting in significant color variations as shown in Figs.~\ref{fig03} and~\ref{fig06}. To address this issue and enhance the color information in the training set, we propose Color Perturbation (CP), which involves randomly exchanging the RGB channels. Specifically, $[R,G,B]$ can be converted to to $[R,B,G]$, $[B,R,G]$, $[G,R,B]$, $[B,G,R]$ and $[G,B,R]$. 
For example, regarding the Color Perturbation (CP) scheme, we employ a random shuffling process on the RGB channels of each image. To elaborate, let's consider an RGB image represented as $I \in \mathbb{R}^{h \times w \times 3}$, where $h$ and $w$ denote the height and width, and 3 signifies the three color channels, namely Red (R), Green (G), and Blue (B). When applying the CP scheme to an image ($\hat{I}$), we perform a random shuffle operation on the vector ($r = [0, 1, 2]$) to obtain $\hat{r}$, which can take one of the following permutations: $\hat{r} = [0, 1, 2]$, $[0, 2, 1]$, $[1, 0, 2]$, $[1, 2, 0]$, $[2, 0, 1]$, or $[2, 1, 0]$. This is mathematically defined as:
\begin{equation} 
\begin{aligned}
\hat{I}[:, :, 0] = I[:, :, \hat{r}[0]]; \\
\hat{I}[:, :, 1] = I[:, :, \hat{r}[1]]; \\
\hat{I}[:, :, 2] = I[:, :, \hat{r}[2]]. 
\end{aligned}
\end{equation}
Fig.~\ref{fig07} shows the data distribution of the original and generated images using t-SNE. The color perturbation introduces diverse information and effectively mitigates overfitting. In the training stage, we randomly combine a color order and the raw image for model training.

\textit{Remark.} Currently, image-level data augmentation is widely used in the computer vision community, with ColorJitter~\cite{Paszke_PyTorch_An_Imperative_2019} being a popular method. This method randomly adjusts the brightness, contrast, and saturation of an image, and has demonstrated its effectiveness in domain generalization classification tasks. However, it is important to note that images captured in urban environments often contain small objects, which differ from the objects used in image classification tasks. For example, in Fig.~\ref{fig07}, some cars are extremely small. If ColorJitter is used for data augmentation, it may result in missing important information about these small objects. We will perform a comparative experiment in the experimental section to further investigate this issue.

While images generated by the CP scheme might not frequently occur in real-world scenes, they exhibit distinctions from the original image in terms of feature distribution while maintaining semantic consistency. Refer to Figure 4 in the main paper for an illustration: the color of these images varies, yet the objects remain consistent with the original image. Thus, training the model with these images can lead to the development of a color-invariant model. In the context of single-domain generalization tasks, images from nighttime scenes differ from those captured during the daytime in terms of color information. In summary, incorporating diverse color information proves beneficial in enabling the model to capture more color-invariant details, a key factor for our task.

Additionally, the proposed CP module can be combined with other image-level augmentations, such as copy-paste, mosaic, and mixup in YOLOv5~\cite{ultralytics_YOLOv5}, to further enhance the model's generalization capability, which will be validated in the experiment.

\subsection{Dual-Style Memory}
In this part, we firstly review Adaptive Instance Normalization (AdaIN)~\cite{DBLP:conf/iccv/HuangB17}, which can transfer the style information from an image to another image. In particular, this work points out that the statistics of feature maps represent the image style information. We define two groups of feature maps as $f, \hat{f}\in \mathbb{R}^{C \times H \times W}$ for two images, where $C$ are the number of channels, and $H$ and $W$ are the hight and the width of feature maps. Here, we assume that the goal is transferring the style information from $\hat{f}$ and $f$, thus we can implement it by:
\begin{equation}
  \begin{aligned}
  &{\rm AdaIN}(f)= \hat{\sigma} \frac{f-\mu}{\sigma}+\hat{\mu},
  \end{aligned}
  \label{eq04}
  \end{equation}
  where  $\mu, \sigma, \hat{\mu}, \hat{\sigma} \in \mathbb{R}^{C}$ (\ie, $\mu=[\mu_1, \cdots,\mu_C]$, $\sigma=[\sigma_1, \cdots,\sigma_C]$, $\hat{\mu}=[\hat{\mu}_1, \cdots,\hat{\mu}_C]$, and $\hat{\sigma}=[\hat{\sigma}_1, \cdots,\hat{\sigma}_C]$) represent the channel-wise mean and standard deviation (\ie, statistics) of $f$. For statistics of the $i$-th channel are presented as:
  \begin{equation}
  \mu_i=\frac{1}{HW}\sum_{h=1}^{H}\sum_{w=1}^{W}f[i, h,w],
    \label{eq05}
  \end{equation}
  \begin{equation}
  \sigma_i=\sqrt{\frac{1}{HW}\sum_{h=1}^{H}\sum_{w=1}^{W}(f[i,h,w]-\mu_i)^2 + \epsilon},
  \label{eq06}
  \end{equation}
  where $\epsilon$ is a constant for numerical stability. Similarly, we can also obtain statistics $(\hat(\mu), \hat{\sigma})$ of $\hat{f}$, which represent the image style information as mentioned before.

To enhance the generalization ability of the model, we generate augmented features to increase the diversity of the abstract style in the feature space. In the single-domain urban object detection task, where the style information is limited, we aim to extract the style information from the entire training data. Additionally, there may be a style discrepancy between the local objects and the background, as shown in Fig.~\ref{fig03}. For example, in Fig.~\ref{fig03}(a), the background is brighter than some cars in the dark region of the image with the red box. Moreover, there exists a style difference between different images based on the local view.


Therefore, in this paper, we propose dual-style memory (DSM) to reach this goal, which saves the style information in a dual-memory. To be specific, we first generate two style memories to save the object style information and the background style information, respectively. Here, we use $\mathbf{M_{obj}}$ and $\mathbf{M_{back}}$ to denote the memory used for saving the style information of the object and the background. We assume that $\mathbf{M_{obj}}$ and $\mathbf{M_{back}}$ have saved some style information. For an input image with $N_{o}$ object(s), its middle feature maps based on convolutional layer (\eg, the feature after each block in ResNet~\cite{DBLP:conf/cvpr/HeZRS16}) are defined as $f$. We split $f$ into different patches according the ground-truth, \eg, $f^{b} \in \mathbb{R}^{C \times A_{b}}$ and $\left \{f^{o1} \in \mathbb{R}^{C \times A_{o1}},\cdots, f^{oN_{o}}\in \mathbb{R}^{C \times A_{oN_{o}}}\right \}$ are used to denote the feature maps of the background and the object set, where $C$ is the number of channels and $A_b$ and $A_{oi}$ is the area of the background and the $i$-th object in the spatial dimension. For the first object, the style information can be represented as $\mu^{o1}=[\mu_1^{o1}, \cdots,\mu_C^{o1}]$ and  $\sigma^{o1}=[\sigma_1^{o1}, \cdots,\sigma_C^{o1}]$. Thus, we can extract its style information of the $i$-th channel as:

  \begin{equation}
  \mu_{i}^{o1}=\frac{1}{A_{o1}}\sum_{a=1}^{A_{o1}}f^{o1}[i, a],
    \label{eq02}
  \end{equation}
  \begin{equation}
  \sigma_i=\sqrt{\frac{1}{A_{o1}}\sum_{a=1}^{A}(f^{o1}[i,a]-\mu_{i}^{o1})^2 + \epsilon}.
  \label{eq03}
  \end{equation}

Similarly, we can compute the style information of the background and all objects, and then save these information into the corresponding style memory. For example, $(\mu^{o1},\sigma^{o1})$ is saved into $\mathbf{M_{obj}}$. Based on the $\mathbf{M_{obj}}$ and $\mathbf{M_{back}}$, we can mine the diverse style information as:

\begin{equation}
  \begin{aligned}
  & \hat{f}^{o1}= \mathbf{M_{back}}[r]_{\sigma} \frac{f^{o1}-\mu^{o1}}{\sigma^{o1}}+\mathbf{M_{back}}[r]_{\mu},
  \end{aligned}
  \label{eq01}
  \end{equation}
where $\mathbf{M_{back}}[r]_{\sigma}$ and $\mathbf{M_{back}}[r]_{\mu}$ is randomly selecting the $r$-th style information in $\mathbf{M_{back}}$. In the meanwhile, we use the same scheme to enhance the style's diversity on all objects and the background. The detailed forward process of the proposed DSM is shown in Alg.~\ref{alg1}.

 \begin{algorithm}[ht]
\caption{\small{The forward process of dual-style memory (DSM).}}~\label{alg1}
\begin{algorithmic}[1]
\STATE {\bf Input:} 
The feature maps $f$. \\
\STATE {\bf Output:} The normalized feature maps $\hat{f}$. \\
\STATE Split $f$ into different patches ($\mathbf{F} = \left \{f^b, f^{o1}, \cdots, f^{oN_{o}}\right \}$) according to the ground-truth.
\FOR {iteration $\in [1,...,N_{o}+1]$}
\STATE Compute the style information for each patch in $\mathbf{F}$ as Eqs.~\ref{eq02} and~\ref{eq03}. \\
\bl{// $|\mathbf{M}|$ is the number of elements in the queue, and $N_m$ is the maximum length of the queue.}
\IF {$|\mathbf{M_{back}}| >= N_m$ or $|\mathbf{M_{obj}}| >= N_m$ }
\STATE Remove the earliest stored style information from the style memory.
\ENDIF
\STATE Save the corresponding style information to the style memory ($\mathbf{M_{back}}$ or $\mathbf{M_{obj}}$).
\STATE Randomly select the style information from the crossed style memory.
\STATE Conduct the adaIN as Eq.~\ref{eq03} to normalize all patches in $\mathbf{F}$.
\ENDFOR
\STATE Splice all patches according the original position.
\end{algorithmic}
\label{al01}
\end{algorithm}

 \textit{Remark.} For the dual-style memory module, we use a fixed-length queue to implement it, which does not require a large amount of memory. As the training set is shuffled at each epoch, the available dual-style memory for a specific sample varies at each epoch, allowing us to extract more diverse information from other samples. Besides, we also conduct the experiment using a shared memory to save all styles and access the style information from the corresponding style memory. This further highlights the importance of these steps in our method.

\begin{figure*}[t]
\centering
\includegraphics[width=14cm]{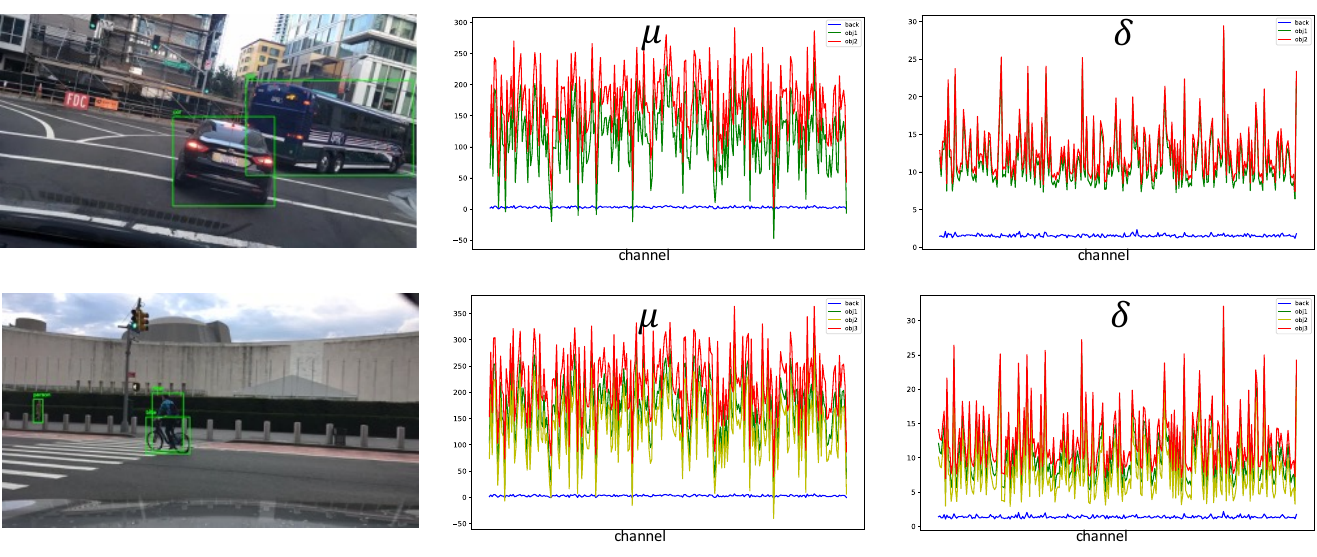}
\caption{An illustration of the statistics of different objects and backgrounds. These statistics (\ie, mean ($\mu$) and variance ($\sigma$)) with 1024-dimension are captured from the first layer of the ResNet-101 pre-trained on ImageNet. In this figure, ``back'' denotes the background, and ``objX'' is the object (\ie, foreground). The first column is the image, and the second and third columns denote the mean and variance, respectively.}
\label{figR01}
\end{figure*}

 Concerning the Dual-Style Memory (DSM), as feature statistics inherently encapsulate style information~\cite{DBLP:conf/iccv/HuangB17}, and the style perspective distinctly highlights the contrast between background and foreground, as illustrated in Fig.~\ref{figR01}.  Specifically, given that feature statistics effectively represent an image's style, we compute the statistics, namely the mean ($\mu$) and variance ($\sigma$), of images from the first layer of ResNet-101, which is pre-trained on ImageNet. According to this figure, it's evident from the figure that there is a noticeable distinction in style information between the foreground and background.
Furthermore, as indicated by the visual statistics, the foreground and background elements from different images also exhibit variations. To address this, we have introduced a dual-style memory that facilitates the generation of diverse samples for model training.
  This memory repository is designed to store the style information corresponding to both foreground and background. During the augmentation process, we randomly and interchangeably draw style information from this dual-memory. This means we can utilize the foreground style for the background or the background style for the foreground. This strategy generates diverse style information for each object and the background.

\section{Explanation of DoubleAUG via Existing Theory}\label{EDET}
In this section, we utilize the domain generalization error bound~\cite{albuquerque2019generalizing} to further demonstrate the effectiveness of our method. In the following part, we firstly review the domain generalization error bound and then analyze our method based on it.

\textbf{Theorem 1}~\cite{albuquerque2019generalizing,DBLP:conf/ijcai/0001LLOQ21} (Domain generalization error bound): Let $\gamma := min_{\pi \in \bigtriangleup_{M}} d_{\mathcal{H}}(\mathcal{P}_{X}^t, \sum_{i=1}^{M}\pi_{i}\mathcal{P}_{X}^i)$ with minimizer $\pi^*$ be the distance of $P_{X}^t$ from the convex hull $\Lambda$, and $P_{X}^* := \sum_{i=1}^{M}\pi_{i}^{*}P_{X}^{i}$ be the best approximator within $\Lambda$. Let $\rho := \sup_{\mathcal{P}_{X}^{'}, \mathcal{P}_{X}^{''}\in \Lambda}d_{\mathcal{H}}(\mathcal{P}_{X}^{'}, \mathcal{P}_{X}^{''})$ be the diameter of $\Lambda$. Then it holds that

\begin{equation}
  \begin{aligned}
\epsilon^{t}(h)\leqslant \sum_{i=1}^{M}\pi_{i}^{*}\epsilon^{i}(h)+\frac{\gamma+\rho}{2}+\lambda_{\mathcal{H}}(\mathcal{P}_X^t, \mathcal{P}_X^*)),
    \end{aligned}
  \label{eq22}
  \end{equation}
where $\lambda_{\mathcal{H}}(\mathcal{P}_X^t, \mathcal{P}_X^*))$ is the ideal joint risk across the target domain and the training domain ($P_X^*$) with the most similar distribution to the target domain.

In Theorem 1, the first item aims to minimize the empirical error on the training set, which can be achieved by the general loss function for object detection. The last item can be treated as a constant. Therefore, we primarily focus on analyzing the second item, which involves $\gamma$ and $\rho$.

Firstly, $\gamma$  represents the discrepancy between the combination of all training domains and the target domain. In the single-domain generalization object detection setting, there is a risk that if the testing domain is far from the training domain in terms of distribution, the model's generalization will be poor for all testing samples. However, our method generates diverse style information based on multiple different distributions, which can be viewed as different domains. Therefore, introducing diverse style information based on CP and DSM can be beneficial in reducing overfitting to the raw single training set and effectively mitigating the aforementioned risk.

Secondly, $\rho$ indicates the maximum distance between different domains. In our method, we extract diverse style information from the training data itself using DSM, while the color perturbation only involves switching the RGB channels. This shows that generating diverse style information in our method does not bring a large domain gap between training samples. Furthermore, we apply DSM to the shallow layer of the neural network (where it mainly focuses on  style information), which also helps to prevent the generation of a large $\rho$ . In summary, our method has an advantage in reducing the generalization error bound from both the $\gamma$ and $\rho$ perspectives in Eq.~\ref{eq22}.

\section{Experiments}\label{s-experiment}
\renewcommand{\cmidrulesep}{0mm} 
\setlength{\aboverulesep}{0mm} 
\setlength{\belowrulesep}{0mm} 
\setlength{\abovetopsep}{0cm}  
\setlength{\belowbottomsep}{0cm}
This part describes the experimental setup and evaluation of the proposed method. Section~\ref{sec:EXP-DS} introduces the datasets and settings used in the experiments. Section~\ref{sec:EXP-CUA} compares the proposed method with state-of-the-art generalizable object detection methods. Ablation studies are conducted in Section~\ref{sec:EXP-SS} to validate the effectiveness of various components in the proposed framework. Lastly, Section~\ref{sec:EXP-FA} further analyzes the properties of the proposed method.

\subsection{Datasets and Experimental Settings}\label{sec:EXP-DS}
\subsubsection{Datasets} 
\textbf{Diverse-Weather}\cite{WuAming_2022_CVPR} is a dataset that includes five scenes with different weather conditions, including daytime-sunny, night-sunny, dusk-rainy, night-rainy, and daytime-foggy. The training set consists of 19,395 images from the daytime-sunny scene, while the testing sets include 8,313 images from the daytime-sunny scene, and 26,158, 3,501, 2,494, and 3,775 images from the night-sunny, dusk-rainy, night-rainy, and daytime-foggy scenes, respectively.

\textbf{SIM10k2Cityscapes.}  is a dataset that combines SIM10k and Cityscapes datasets. \textit{SIM10k} \cite{DBLP:conf/icra/Johnson-Roberson17}  consists of 10,000 images rendered by the Grand Theft Auto (GTAV) gaming engine, with bounding boxes of 58,701 cars provided in the 10,000 training images. In our experiments, we randomly selected 9,000 images for training and 1,000 for testing. 
\textit{Cityscapes \& Foggy Cityscapes \& Rain Cityscapes}.  Cityscapes \cite{DBLP:conf/cvpr/CordtsORREBFRS16} is a traffic scene dataset for driving scenarios. The images are captured by a car-mounted video camera. It has 2,975 images in the training set, and 500 images in the validation set. We follow \cite{DBLP:conf/cvpr/Chen0SDG18,DBLP:conf/eccv/HsuTLY20} to use the validation set as the target domain to test our method. Foggy Cityscapes \cite{SDV18} is a fog-rendered Cityscapes dataset, it has 8,877 images in the training set, and 1,473 images in the validation set, the same as Cityscapes we use the validation set as the target domain. Rain Cityscapes \cite{Hu_2019_CVPR} is a rain-rendered Cityscapes dataset, it has 9,432 training images and 1,188 testing images, the same as Cityscapes and Foggy Cityscapes, we only use the validation set as the target domain. There are 8 categories with instance labels in all Cityscapes \& Foggy Cityscapes \& Rain Cityscapes, but the only car is used in this experiment since the only one is annotated in SIM 10k. Note that the Cityscapes \& Foggy Cityscapes \& Rain Cityscapes dataset is not dedicated to the detection, thus we take the tightest rectangle of its instance masks as ground-truth bounding boxes.

\subsubsection{Implementation Details} 
We conduct all experiments using PyTorch 1.8.2 \cite{Paszke_PyTorch_An_Imperative_2019} with Detectron2 \cite{wu2019detectron2} library. For all experiments, we both use Faster R-CNN \cite{DBLP:conf/nips/RenHGS15} and YOLOv5 \cite{ultralytics_YOLOv5} as our base detectors. In particular, YOLOv5s (14.12MB) is a smaller model than Faster R-CNN (232.2MB). We use mAP (\%) as the main evaluation metric when IOU = 0.5.
For Faster R-CNN \cite{DBLP:conf/nips/RenHGS15}, ResNet-101 \cite{DBLP:conf/cvpr/HeZRS16} is taken as the backbone, and we use the weights pre-trained on COCO \cite{DBLP:conf/eccv/LinMBHPRDZ14} in initialization (provided by Detectron2 \cite{wu2019detectron2}), all models are trained on 2 GPUs using SGD with a mini-batch size of 4, the momentum is 0.9, the max iterations is 100,000, and the learning rate is 0.001, we also apply warmup by 5,000 iterations. 
For YOLOv5 \cite{ultralytics_YOLOv5} we choose YOLOv5s as our baseline, and we use the weights pre-trained on COCO \cite{DBLP:conf/eccv/LinMBHPRDZ14} in initialization (provided by YOLOv5 \cite{ultralytics_YOLOv5}), all models are trained on 2 GPUs using SGD with a mini-batch size of 44, the momentum is 0.843, the max epoch is 200, and the learning rate is 0.0032. 
 Note that we obtain the final model from the last epoch for all experiments. Unless otherwise specified, Faster R-CNN is used as the default baseline in all experiments.

\subsection{Comparison with State-of-the-art Methods}\label{sec:EXP-CUA}

\begin{table}[htbp]
  \centering
  \caption{Experimental results (\%) on the Diverse-Weather dataset. All methods are trained on Daytime-Sunny, and tested on Night-Sunny, Dusk-Rainy, Night-Rainy, and Daytime-Foggy. Note that All these methods with the available code provided by its authors are run on the Diverse-Weather dataset.}
    \begin{tabular}{l|cccc|c}
    \toprule
    \multicolumn{1}{c|}{Method} & Night-Sunny & Dusk-rainy & Night-Rainy & Daytime-Foggy & mAP \\
    \midrule
    SW\cite{DBLP:conf/iccv/PanZST019}   & 39.63  & 33.85  & 16.07  & 36.06  & 31.40  \\
    IBN-Net\cite{DBLP:conf/eccv/PanLST18} & 40.42  & 36.80  & 18.06  & 35.84  & 32.78  \\
    IterNorm\cite{DBLP:conf/cvpr/HuangZ00019} & 40.14  & 34.25  & 14.58  & 34.28  & 30.81  \\
   \midrule
    CycConf\cite{DBLP:conf/iccv/0066HLY0GD21}  & 45.25 & 37.00  & 16.63  & 38.06  & 34.23  \\
    CycConf+Ours & 45.04  & 42.12  & 18.50  & 38.95  & 36.15  \\
    \midrule
    CDSD\cite{WuAming_2022_CVPR}  & 37.38  & 27.16  & 12.94  & 33.33  & 27.70  \\
    CDSD+Ours & 36.43  & 31.91  & 15.62  & 35.74  & 29.93  \\
    \midrule
    Faster R-CNN\cite{DBLP:conf/nips/RenHGS15} & 46.19  & 38.01  & 16.23  & 38.14  & 34.64  \\
    Faster R-CNN+Ours & 47.13  & 41.48  & 22.17  & 39.01  & 37.45  \\
    \bottomrule
    \end{tabular}%
  \label{tab01}%
\end{table}%

In this part, we perform the experiment to compare our method with some SOTA methods, including SW~\cite{DBLP:conf/iccv/PanZST019}, IBN-Net~\cite{DBLP:conf/eccv/PanLST18}, IterNorm~\cite{DBLP:conf/cvpr/HuangZ00019} ISW~\cite{DBLP:conf/cvpr/ChoiJYKKC21}, CycConf~\cite{DBLP:conf/iccv/0066HLY0GD21} and CDSD~\cite{WuAming_2022_CVPR}. Particularly, CycConf and CDSD are designed for the generalizable objection detection task. CycConf improves the generalization on the out-of-distribution dataset via a novel self-training method. CDSD is a recent method for the single-domain generalized object detection in the traffic scene, which aims to extract the domain-invariant feature by the cyclic-disentangled self-distillation. 
For all methods, we run the experiment based the available code provided by authors.
Table~\ref{tab01} is the result of the Diverse-Weather dataset. In this experiment, we use the same dataset and dataset setting as CDSD. As observed in this table, our method outperforms all methods based on Faster R-CNN. 
Furthermore, when applying our method to CycConf and CDSD, it can also further enhance the generalizable capability. Note that, the result of CDSD differs from the result in \cite{WuAming_2022_CVPR}, because the dataset provided by \cite{WuAming_2022_CVPR} is not divided into the training set and testing set. Hence, although we split the dataset according to the number of samples the same as \cite{WuAming_2022_CVPR}, the training samples and testing samples are not completely consistent with \cite{WuAming_2022_CVPR}.


Moreover, we also conduct the comparison in the case from SIM10k to Cityscapes, as reported in Table~\ref{tab02}. Since SIM10k is from the game, which is the virtual dataset, it has a large difference when compared to Cityscapes collected from the real-world city scenario. Similar to the above analysis in Table~\ref{tab01}, ``Faster R-CNN'' and ``Faster R-CNN+Ours'' exists an obvious difference in all domains.


\begin{table}[htbp]
  \centering
  \caption{Experimental results (\%) of domain generalization from SIM10K to Cityscapes. Raw, Rain and Foggy are the different domains of Cityscapes. Note that we run the all methods on SIM10k2Cityscapes based on the available code provided by their authors.}
    \begin{tabular}{l|ccc|c}
    \toprule
    \multicolumn{1}{c|}{Method} & Raw   & Rain  & Foggy & mAP \\
    \midrule
    SW\cite{DBLP:conf/iccv/PanZST019}     & 44.23  & 30.01  & 34.03  & 36.09  \\
    IBN-Net\cite{DBLP:conf/eccv/PanLST18} & 49.47  & 36.66  & 38.55  & 41.56  \\
    IterNorm\cite{DBLP:conf/cvpr/HuangZ00019} & 39.17  & 21.51  & 30.70  & 30.46  \\
    \midrule
    CycConf\cite{DBLP:conf/iccv/0066HLY0GD21} & 52.54  & 42.97  & 39.25  & 44.92  \\
    CycConf+Ours & 60.72  & 56.41  & 50.92  & 56.02  \\
    \midrule
    CDSD\cite{WuAming_2022_CVPR}  & 33.47  & 14.20  & 17.40  & 21.69  \\
    CDSD+Ours & 40.23  & 28.82  & 27.13  & 32.06  \\
    \midrule
    Faster R-CNN\cite{DBLP:conf/nips/RenHGS15} & 51.21  & 36.73  & 35.16  & 41.03  \\
    Faster R-CNN+Ours & 61.66  & 56.75  & 50.74  & 56.38  \\
    \bottomrule
    \end{tabular}%
  \label{tab02}%
\end{table}%

\subsection{Ablation Studies} \label{sec:EXP-SS}

In this part, we perform the experiment to sufficiently validate the effectiveness of each module in the proposed DoubleAUG on the Diverse-Weather and SIM10k2Cityscapes datasets. The experimental results are listed in Tables~\ref{tab03} and~\ref{tab04}. As seen in these two tables, both the proposed color perturbation (CP) and dual-style memory (DSM) can improve the model's generalization on two datasets. For example, on Diverse-Weather, the CP and DSM outperform the baseline by $+1.18\%$ (35.82 vs. 34.64) and $+2.12\%$ (36.76 vs. 34.64), respectively, which confirms the efficacy of these proposed modules. Furthermore, better performance can be obtained when combining the CP and DSM together. In addition, we also observe that the improvement is significant on SIM10k2Cityscapes when using our method, which is because of the large domain gap between the virtual data (SIM10K) and the real-world data (Cityscapes). Therefore, our method can achieve a great performance improvement when the unseen domain is obviously different from the training set.
\begin{table}[htbp]
  \centering
  \caption{Evaluation of different moudles in our method on Diverse-Weather.}
    \begin{tabular}{l|cccc|c}
    \toprule
    \multicolumn{1}{c|}{Method} & Daytime-Foggy & Dusk-rainy & Night-Rainy & Night-Sunny & mAP \\
    \midrule
    Baseline & 38.14 & 38.01 & 16.23 & 46.19 & 34.64 \\
    Baseline+CP & 38.95 & 37.68 & 19.38 & 47.26 & 35.82 \\
    Baseline+DSM & 39.30 & 40.71 & 20.83 & 46.21 & 36.76 \\
    Baseline+CP+DSM & 39.01 & 41.48 & 22.17 & 47.13 & 37.45 \\
    \bottomrule
    \end{tabular}%
  \label{tab03}%
\end{table}%

\begin{table}[htbp]
  \centering
\caption{Evaluation of different moudles in our method on SIM10k2Cityscapes.}
    \begin{tabular}{l|ccc|c}
    \toprule
    \multicolumn{1}{c|}{Method} & Raw   & Rain  & Foggy & mAP \\
    \midrule
    Baseline & 51.21 & 36.73 & 35.16 & 41.03 \\
    Baseline + CP & 57.42 & 46.77 & 44.23 & 49.48 \\
    Baseline + DSM & 59.60 & 55.48 & 48.98 & 54.69 \\
    Baseline + CP +DSM & 61.66 & 56.75 & 50.74 & 56.38 \\
    \bottomrule
    \end{tabular}%
  \label{tab04}%
\end{table}%

\subsection{Further Analysis}\label{sec:EXP-FA}
\begin{figure*}[t]
\centering
\includegraphics[width=12cm]{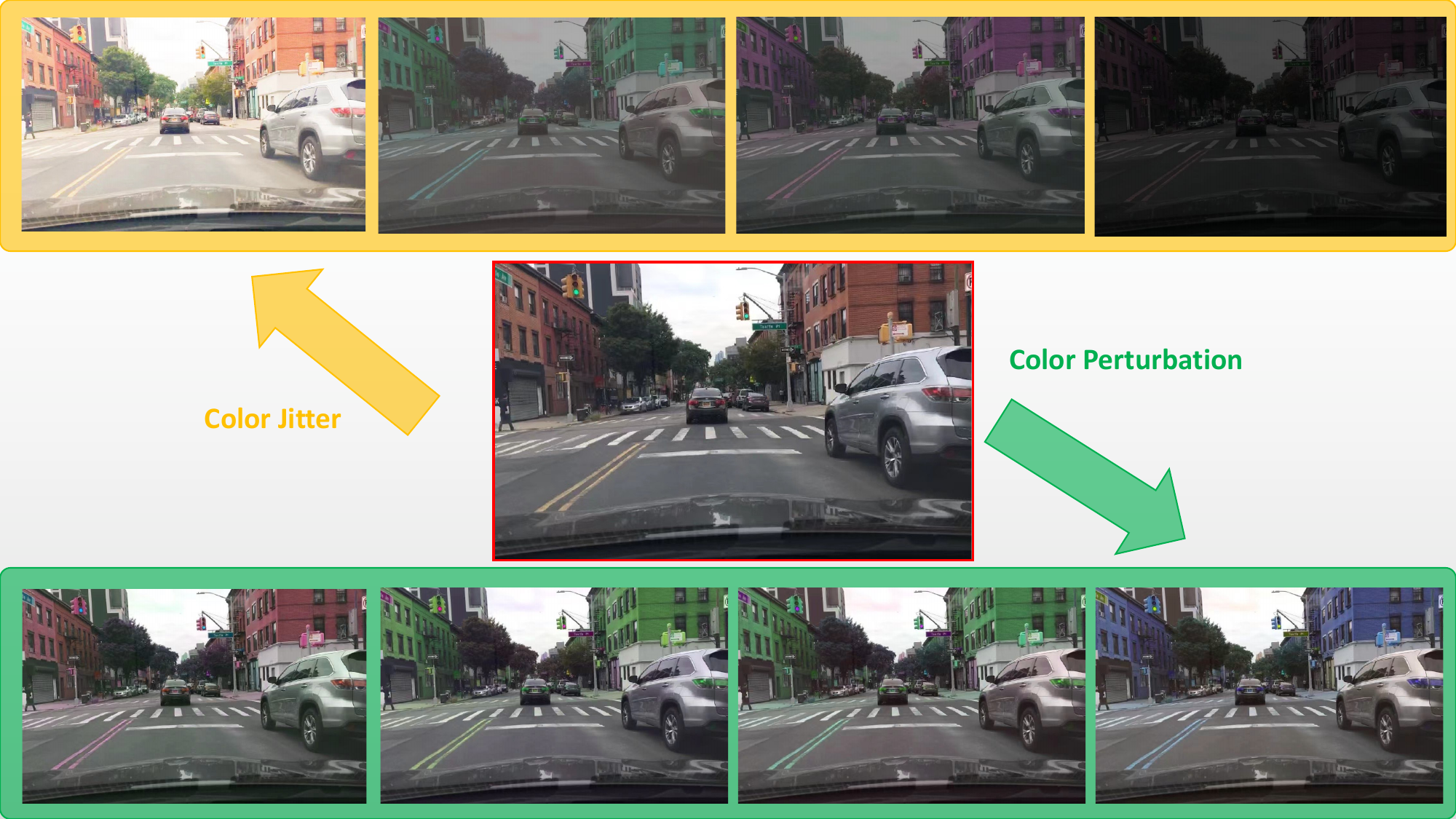}
\caption{The visual comparison between color jitter and our color perturbation.}
\label{fig05} \
\end{figure*}

\textbf{Comparison between the proposed CP and the ColorJitter}. ColorJitter is a type of image data augmentation where we randomly change the brightness, contrast and saturation of an image, and it has been widely used in computer vision. In this experiment, we compare the proposed color perturbation with it on Diverse-Weather and SIM10k2Cityscapes. The experimental results are shown in Tables~\ref{tab09} and~\ref{tab10}. As observed in these tables, the proposed color perturbation can achieve better performance than ColorJitter, \eg, the performance can be increased by $+0.48\%$ (37.45 vs. 36.97) and $+1.21\%$ (56.38 vs. 55.17) on Diverse-Weather and SIM10k2Cityscapes, respectively. The main reason is that these small objects in the urban-scene images could be blurry when using ColorJitter, as illustrated in Fig.~\ref{fig05}.

\begin{table}[htbp]
  \centering
  \caption{Comparison between the proposed Color Perturbation (CP) and ColorJitter (CJ) on Diverse-Weather.}
    \begin{tabular}{l|cccc|c}
    \toprule
    \multicolumn{1}{c|}{Method} & Daytime-Foggy & Dusk-rainy & Night-Rainy & Night-Sunny & mAP \\
    \midrule
    Baseline+DSM+CJ & 39.82 & 40.77 & 20.56 & 46.74 & 36.97  \\
    Baseline+DSM+CP (Ours) & 39.01 & 41.48 & 22.17 & 47.13 & 37.45  \\
    \bottomrule
    \end{tabular}%
  \label{tab09}%
\end{table}%

\begin{table}[htbp]
  \centering
  \caption{Comparison between the proposed Color Perturbation (CP) and ColorJitter (CJ) on SIM10k2Cityscapes.}
    \begin{tabular}{l|ccc|c}
    \toprule
    \multicolumn{1}{c|}{Method} & Raw   & Rain  & Foggy & mAP \\
    \midrule
    Baseline+DSM+CJ & 60.73 & 53.27 & 51.52 & 55.17 \\
    Baseline+DSM+CP (Ours) & 61.66 & 56.75 & 50.74 & 56.38 \\
    \bottomrule
    \end{tabular}%
  \label{tab10}%
\end{table}%

\begin{figure*}[t]
\centering
\includegraphics[width=12cm]{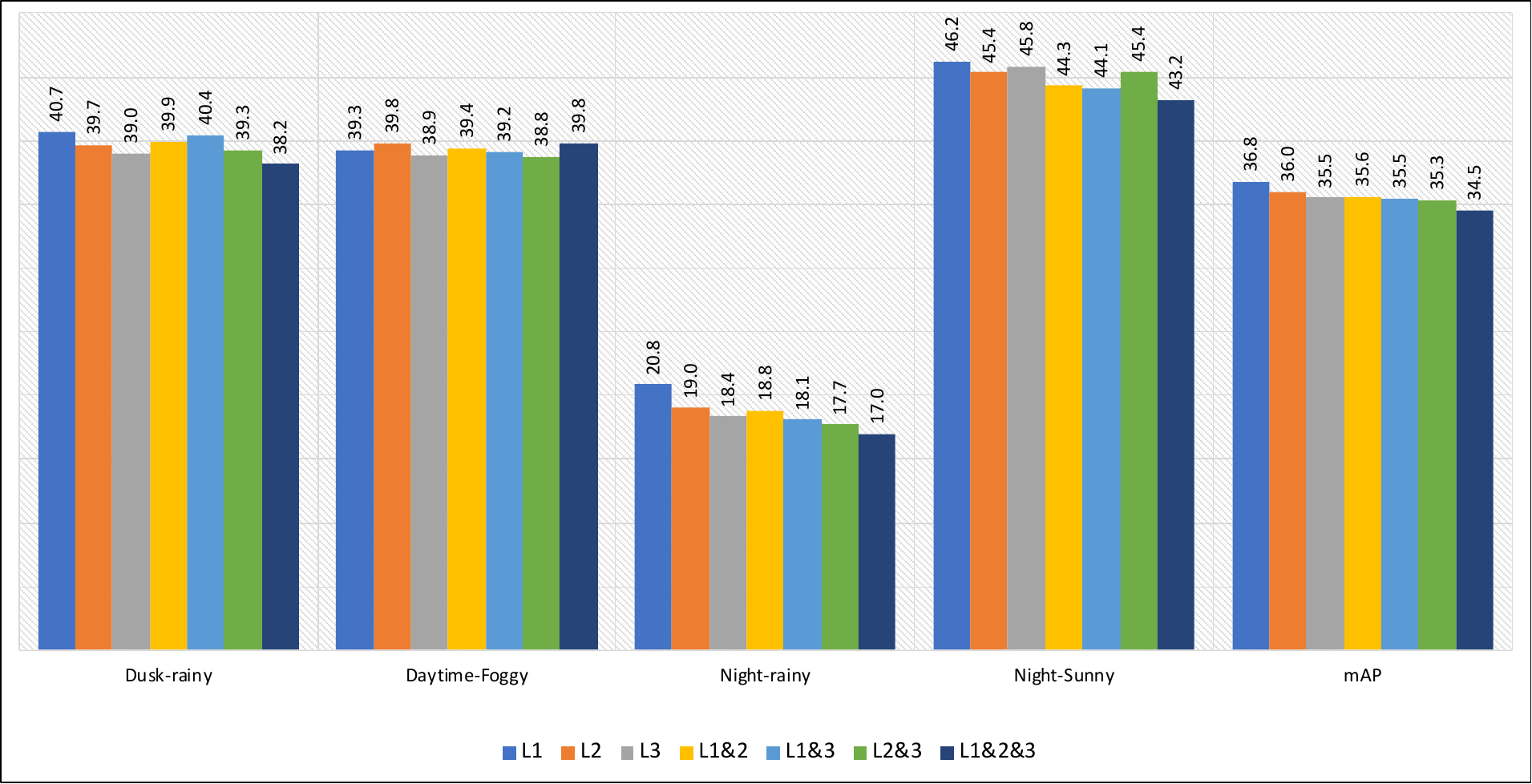}
\caption{Experimental results of the style memory used in different layers on Diverse-Weather. It is worth noting that ``L1'' denotes using the DSM after the first block, and ``L1\&2'' indicates that using the DSM after the first and second blocks simultaneously.}
\label{fig01}
\end{figure*}

\textbf{Evaluation on style memory used in different layers.} 
In this experiment, we report the experimental results when using the proposed dual-style memory (DSM) in different layers, as given in Fig.~\ref{fig01}. As known, the ResNet consists of four blocks, thus we can use the DSM after each block. As a whole, we can obviously find that when using the DSM after the first block can produce the best result. The information from the shallow layer of the neural network denotes the color, texture, and so on, which can be viewed as the style information, while the information from the deep layer of the neural network means the semantic information. Hence, using the proposed DSM to enrich the style in the shallow layer is reasonable.

\textbf{Further evaluation on the DSM.} We further evaluate the necessity of these components in the DSM, as reported in Table~\ref{tab05}. In this experiment, we choose style information for the object (background) from the object (background) style (\ie, ``no-exchange'' in Table~\ref{tab05}), and select the style information for the object (background) from the background (object) memory (\ie, ``exchange'' in Table~\ref{tab05}). As seen in Table~\ref{tab05}, the crossed selection is better than the corresponding selection. In addition, we also perform the experiment using one style memory for saving both object and background styles. As seen in Table~\ref{tab05}, it is effective for using two independent memories for saving object and background respectively.

\begin{table}[htbp]
  \centering
  \caption{Further evaluation for the DSM on Diverse-Weather. ``no-exchange'' is selecting style information for the object (background) from the object (background) style, and ``exchange'' is selecting the style information for the object (background) from the background (object) memory in the top. In the bottom, ``one memory'' is using one style memory for saving both object and background styles, and ``divided memory'' is using two independent memories for saving object and background styles, respectively.}
    \begin{tabular}{l|cccc|c}
    \toprule
    \multicolumn{1}{c|}{Method} & Daytime-Foggy & Dusk-rainy & Night-rainy & Night-Sunny & mAP \\
    \midrule
    no-exchange & 38.97 & 41.09 & 19.97 & 45.31 & 36.34 \\
    exchange (ours) & 39.30 & 40.71 & 20.83 & 46.21 & 36.76 \\
    \midrule
    \midrule
    one memory & 39.30 & 39.91 & 20.23 & 45.36 & 36.20 \\
    divided memory(ours) & 39.30 & 40.71 & 20.83 & 46.21 & 36.76 \\
    \bottomrule
    \end{tabular}%
  \label{tab05}%
\end{table}%

\begin{figure*}[t]
\centering
\includegraphics[width=12cm]{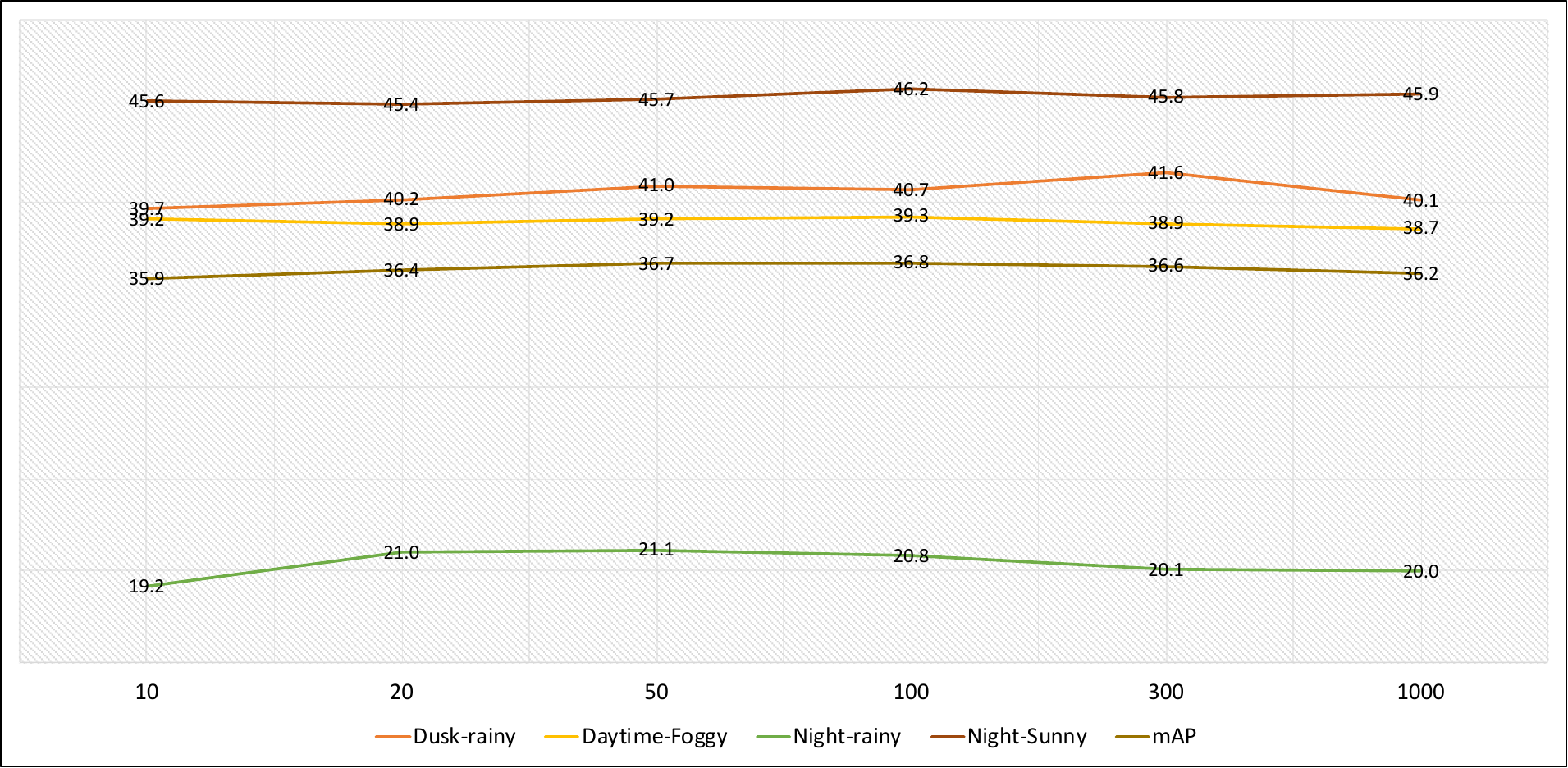}
\caption{Experimental results of the DSM with different memory sizes on Diverse-Weather.}
\label{fig02}
\end{figure*}
\textbf{Experimental results of the DSM with different memory sizes.} We conduct the experiment to observe the influence of different memory sizes in the proposed dual-style memory. As seen in Fig.~\ref{fig02}, we can obtain the best result when the memory size is set as 100. We use the setting in all experiments. 

\textbf{Comparison between the proposed DSM and MixStyle.} MixStyle~\cite{DBLP:conf/iclr/ZhouY0X21} is an augmentation method by mixing the style information of these images in a batch. Since our DSM does not introduce the extra information (\ie, only mining the style information from the training set), it is fair to compare them. The experimental results are listed in Table~\ref{tab07}. We can observe that the DSM outperforms the MixStyle by $0.92$ (36.76 vs. 35.84) on the Diverse-Weather dataset.
\begin{table}[htbp]
  \centering
  \caption{Comparison between the proposed DSM and MixStyle on Diverse-Weather.}
    \begin{tabular}{c|cccc|c}
    \toprule
    \multicolumn{1}{c|}{Method} & Daytime-Foggy & Dusk-rainy & Night-Rainy & Night-Sunny & mAP \\
    \midrule
    Baseline & 38.14 & 38.01 & 16.23 & 46.19 & 34.64 \\
    +MixStyle & 38.99 & 39.50 & 17.61 & 47.26 & 35.84 \\
    +DSM   & 39.30 & 40.71 & 20.83 & 46.21 & 36.76 \\
    \bottomrule
    \end{tabular}%
  \label{tab07}%
\end{table}%

\textbf{Evaluation of the stability of the proposed method.} We conduct five experiments with different random seeds to show the stability of the proposed method, as reported in Table~\ref{tab08}. As seen, the STD of the baseline is 0.25, while our method is 0.08. This result shows our method is stable.

\begin{table}[htbp]
  \centering
  \caption{Evaluation of the stability of the proposed method on Diverse-Weather. In this table, ``AVG'' means the averaged result five times, and ``STD'' is the corresponding standard deviation.}
    \begin{tabular}{c|c|cccc|c}
    \toprule
    Method & \multicolumn{1}{c|}{Seed} & Daytime-Foggy & Dusk-rainy & Night-Rainy & Night-Sunny & mAP \\
    \midrule
    \multirow{7}[4]{*}{Baseline} & \multicolumn{1}{c|}{1} & 38.14 & 38.01 & 16.23 & 46.19 & 34.64 \\
          & 2     & 37.27 & 38.79 & 16.28 & 46.76 & 34.78 \\
          & 3     & 38.04 & 38.80 & 17.28 & 46.75 & 35.22 \\
          & 4     & 37.04 & 38.73 & 15.98 & 46.65 & 34.60 \\
          & 5     & 37.45 & 38.43 & 16.94 & 46.70 & 34.88 \\
\cmidrule{2-7}          & AVG   & 37.59 & 38.55 & 16.54 & 46.61 & 34.82 \\
          & STD   & 0.48  & 0.34  & 0.54  & 0.24  & 0.25 \\
    \midrule
    \multirow{7}[4]{*}{Ours} & \multicolumn{1}{c|}{1} & 39.01 & 41.48 & 22.17 & 47.13 & 37.45 \\
          & 2     & 39.46 & 40.83 & 21.55 & 47.39 & 37.31  \\
          & 3     & 39.45 & 41.08 & 21.57 & 47.58 & 37.42  \\
          & 4     & 39.35 & 41.68 & 21.79 & 47.34 & 37.54  \\
          & 5     & 39.26 & 41.85 & 21.14 & 47.61 & 37.47  \\
\cmidrule{2-7}          & AVG   & 39.31 & 41.38 & 21.64 & 47.41 & 37.44 \\
          & STD   & 0.18  & 0.42  & 0.38  & 0.20  & 0.08 \\
    \bottomrule
    \end{tabular}%
  \label{tab08}%
\end{table}%

\textbf{Experimental results on the source domain.} We here show the result on the source domain in Table~\ref{tab06}. We find that our method decreases the performance on the source domain when compared with the baseline, which can be explained by the fact that our method can effectively reduce the overfitting risk in the training stage. Hence our DoubleAUG has the ability to generalize well to unseen domains.
\begin{table}[htbp]
  \centering
  \caption{Experimental results on the source domain in the Diverse-Weather and SIM10k2Cityscapes tasks.}
    \begin{tabular}{l|c|c}
    \toprule
    \multicolumn{1}{c|}{Method} & Daytime-sunny & SIM10K \\
    \midrule
    Baseline & 64.18 & 89.15 \\
    Baseline + DoubleAUG (Ours) & 61.05 & 87.32 \\
    \bottomrule
    \end{tabular}%
  \label{tab06}%
\end{table}%

\textbf{Experimental results of different modules based on YOLOv5s.}  
We conduct the experiment based on YOLOv5s, which is a small model than the Faster R-CNN based on ResNet-101. Besides, unlike the two-stage Faster R-CNN, it is a one-stage object detection method. We report the experimental results in Tables~\ref{tab11} and~\ref{tab12}. It is worth noting that we first perform the experiment based on the clean YOLOv5s (\ie, removing these augmentation schemes including copy-past, mosaic and mixup.)  As displayed in these two tables, each module in our method is effective, especially on SIM10k2Cityscapes, our method can achieve significant improvement. In addition, we also conduct the experiment based on the whole YOLOv5s (\ie, using all raw augmentation schemes in YOLOv5s). As seen in Tables~\ref{tab11} and~\ref{tab12}, our method also can achieve an obvious improvement.  
\begin{table}[htbp]
  \centering
  \caption{Experimental results of different modules based on YOLOv5s on Diverse-Weather. ``w/o AUG'' means that we remove these augmentation schemes including copy-past, mosaic and mixup.}
  \resizebox{\textwidth}{!}
  {
    \begin{tabular}{l|cccc|c}
    \toprule
    \multicolumn{1}{c|}{Method} & Daytime-Foggy & Dusk-rainy & Night-Rainy & Night-Sunny & mAP \\
    \midrule
    YOLOv5s w/o AUG & 25.5  & 33.5  & 10.4  & 38.1  & 26.9  \\
    YOLOv5s w/o AUG+ CP & 28.2  & 34.5  & 12.1  & 38.6  & 28.4  \\
    YOLOv5s w/o AUG+ CP+DSM & 31.7  & 33.4  & 16.2  & 38.9  & 30.1  \\
    \midrule
    YOLOv5s & 28.4  & 36.8  & 14.5  & 39.5  & 29.8  \\
    YOLOv5s + CP & 30.7  & 37.1  & 15.9  & 39.1  & 30.7  \\
    YOLOv5s + CP+DSM & 32.1  & 37.5  & 17.6  & 39.2  & 31.6  \\
    \bottomrule
    \end{tabular}%
    }
  \label{tab11}%
\end{table}%

\begin{table}[htbp]
  \centering
  \caption{Experimental results of different modules based on YOLOv5s on SIM10k2Cityscapes. `w/o AUG'' means that we remove these augmentation schemes including copy-past, mosaic and mixup.}
    \begin{tabular}{l|ccc|c}
    \toprule
    \multicolumn{1}{c|}{Method} & Raw   & Rain  & Foggy & mAP \\
    \midrule
    YOLOv5s w/o AUG & 24.7  & 11.8  & 13.9  & 16.8  \\
    YOLOv5s w/o AUG+ CP & 28.6  & 18.2  & 18.5  & 21.8  \\
    YOLOv5s w/o AUG+ CP+DSM & 34.7  & 21.2  & 20.9  & 25.6  \\
    \midrule
    YOLOv5s & 28.5  & 25.5  & 25.1  & 26.4  \\
    YOLOv5s + CP & 31.5  & 26.9  & 27.4  & 28.6  \\
    YOLOv5s + CP+DSM & 37.8  & 26.6  & 28.1  & 30.8  \\
    \bottomrule
    \end{tabular}%
  \label{tab12}%
\end{table}%

\textbf{Comparison our method with two-stage scheme.}
To further demonstrate the effectiveness of our approach, we initially leverage the recent method proposed in \cite{fei2023generative}, which was published in CVPR 2023, to enhance image quality. Subsequently, we conduct the detection process in accordance with your recommendation. The results of these experiments are presented in Tabs.~\ref{tabR01} and~\ref{tabR02}.
As observed in these tables, our method clearly outperforms the two-stage scheme.

\renewcommand{\cmidrulesep}{0mm} 
\setlength{\aboverulesep}{0mm} 
\setlength{\belowrulesep}{0mm} 
\setlength{\abovetopsep}{0cm}  
\setlength{\belowbottomsep}{0cm}

\begin{table}[htbp]
  \centering
  \caption{Comparison our method with the two-stage scheme on Diverse-Weather.}
    \begin{tabular}{l|cccc|C{0.5cm}}
    \toprule
    \multicolumn{1}{c|}{Method} & Night-Sunny & Dusk-rainy & Night-Rainy & Daytime-Foggy & mAP \\
    \midrule
    Baseline & 38.14 & 38.01 & 16.23 & 46.19 & 34.64 \\
    Two-stage & 37.89 & 37.51 & 18.45 & 48.93 & 35.70 \\
    Ours  & 39.01 & 41.48 & 22.17 & 47.13 & 37.45 \\
    \bottomrule
    \end{tabular}%
  \label{tabR01}%
\end{table}%

\begin{table}[htbp]
  \centering
  \caption{Comparison our method with the two-stage scheme on SIM10k2Cityscapes.}
    \begin{tabular}{l|ccc|c}
    \toprule
    \multicolumn{1}{c|}{Method} & Raw   & Rain  & Foggy & mAP \\
    \midrule
    Basline & 51.20 & 36.74 & 35.16 & 41.03 \\
    Two-stage & 52.30 & 37.45 & 34.17 & 41.31 \\
    Ours  & 61.66 & 56.75 & 50.74 & 56.38 \\
    \bottomrule
    \end{tabular}%
  \label{tabR02}%
\end{table}%

\section{Conclusion}\label{s-conclusion}

In this paper, we propose a simple yet effective approach, DoubleAUG, to address the single-domain generalization problem in object detection tasks in urban scenes. Our approach comprises two modules: image-level color perturbation (CP) and feature-level dual-style memory (DSM). The CP module randomly shuffles the RGB channels to generate diverse color information, while the DSM module utilizes object and background style memories to save and extract diverse style information across the entire dataset. We conduct experiments on multiple tasks and settings to demonstrate the effectiveness of our proposed method. Additionally, we employ existing domain generalization theory to analyze the properties of our approach.

As noted in our experiment, our method effectively mitigates overfitting to the source domain, as demonstrated in Tab.~\ref{tab06}. Consequently, when our model is employed in scenarios resembling the training domain, its performance may exhibit a decrease compared to the baseline. This situation is particularly challenging in real-world applications, as distinguishing the domain of origin for a given image is often not possible.
In our future work, we intend to enhance the model's performance on the source domain while simultaneously preserving its generalization capabilities to other unseen domains.


%
%

{\small
\bibliographystyle{acm}
\bibliography{egbib}
}
\end{document}